\documentclass[runningheads]{llncs}
\usepackage{graphicx}
\usepackage{algorithm}
\usepackage{algpseudocode}
\usepackage{amsmath}
\usepackage{array}
\usepackage[hidelinks]{hyperref}
\usepackage[affil-it]{authblk}
\usepackage{todonotes}
\usepackage{enumitem}

\begin{document}
\title{Neural Decompiling of Tracr Transformers}

\authorrunning{Hannes Thurnherr and Kaspar Riesen}

\author{Hannes Thurnherr\orcidID{0009-0001-4554-3899} \and
Kaspar Riesen\orcidID{0000-0002-9145-3157}}

\institute{Institute of Computer Science, University of Bern, 3012 Bern, Switzerland\\
\email{hannes.thurnherr@students.unibe.ch, kaspar.riesen@unibe.ch}}

\maketitle             
\begin{abstract}
Recently, the transformer architecture has enabled substantial progress in many areas of pattern recognition and machine learning. However, as with other neural network models, there is currently no general method available to explain their inner workings. The present paper represents a first step towards this direction. We utilize \textit{Transformer Compiler for RASP} (Tracr) to generate a large dataset of pairs of transformer weights and corresponding RASP programs. Based on this dataset, we then build and train a model, with the aim of recovering the RASP code from the compiled model. We demonstrate that the simple form of Tracr compiled transformer weights is interpretable for such a decompiler model. In an empirical evaluation, our model achieves exact reproductions on more than 30\% of the test objects, while the remaining 70\% can generally be reproduced with only few errors. Additionally, more than 70\% of the programs, produced by our model, are functionally equivalent to the ground truth, and therefore a valid decompilation of the Tracr compiled transformer weights.

\end{abstract}
\section{Introduction}

The \textit{transformer architecture}~\cite{vaswani2017attention} is a type of neural network that was originally designed for machine translation but is now also successfully used for modelling many different pattern recognition tasks. This notably includes advanced language understanding~\cite{radford2018improving} but also the modelling of non-sequential data such as images~\cite{dosovitskiy2020image}. The transformer architecture differentiates itself from other architectures by using a mixture of self-attention and MLP-layers. It can be used as both a decoder-only variant or in an encoder-decoder configuration where the encoder augments the decoder forward pass using cross-attention.

\textit{Interpretability} is a subfield of machine learning that is concerned with the problem of understanding the internals of neural networks~\cite{zhang2021survey}. Especially the interpretability of transformer-based models has seen growing attention in the past few years. While some progress has been made, e.g.~\cite{bricken2023towards,olsson2022context,bills2023language}, transformer neural networks are still largely regarded as black boxes. That is, interpretation relies on manual work which makes it hard to scale (for instance, manually translating billions of model weights and activations into human-readable descriptions is hardly feasible -- even if one knew how to do so). The fact that current state-of-the-art systems are not interpretable, means that they could have unacceptable failure modes, which may only be discovered after deployment.

Attempts to automate at least parts of the complete interpretation pipeline have been made~\cite{ahmad2024causal,conmy2024towards}. However, most of these approaches only focus on the analysis of a subcomponent of the network. 
In the present paper, we propose an end-to-end framework, where a decompiler model is responsible for the entire interpretation process of a complete neural network (taking all of its parameters into account). In particular, we investigate whether it is possible to train a deep-learning model to automate the translation of a set of compiled transformer weights, into \textit{RASP}\footnote{\textit{Restricted Access Sequence Processing Language}} code~\cite{weiss2021thinking}. RASP is a programming language that consists of functions that correspond to various components of transformers and is designed to be computationally equivalent to transformers. RASP code could thus open up a way of translating transformation processes into a form that is more readable for humans.

\begin{figure}
    \centering
    \includegraphics[width=0.6\linewidth]{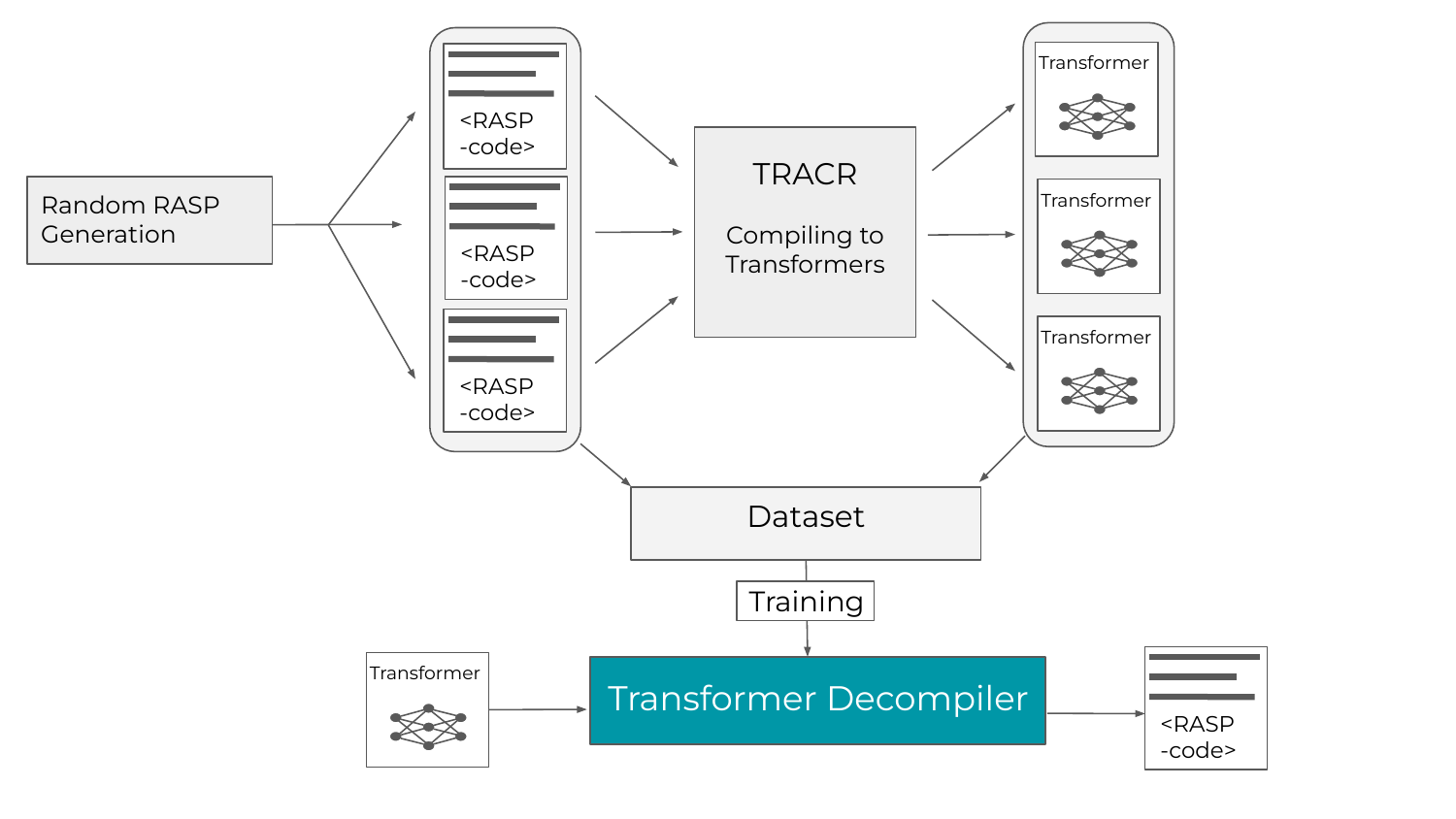}
    \caption{The core concept of the present method is based on the idea that Tracr can be used to generate training data for an automated end-to-end interpretability system.}
    \label{fig:Trade_concept}
\end{figure}

One of the major challenges of the present research is the generation of a large amount of training data. To date,  only a few examples of pairs of network weights and corresponding descriptions are available (from which a neural network could learn to translate between the two). To address this challenge, we use the \textit{Transformer Compiler for Rasp}~(Tracr)~\cite{lindner2024Tracr} to generate a novel set of training data. In particular, we use Tracr to produce a large set of pairs of transformer weights and corresponding RASP programs. We hypothesize that using these pairs to train a \textit{Transformer Decompiler Model} could represent a first step towards a holistic and fully automated solution for transformer interpretability. In Fig.~\ref{fig:Trade_concept}, the proposed approach of the present paper is visually summarized.

The remainder of the paper is organized as follows. Next, in Section~\ref{dataset}, we describe in detail how the dataset, used to train our model, is engineered. Then, in Section~\ref{method}, we outline the novel method for processing the RASP transformer weight pairs and training the decompiler model. In Section~\ref{exp_eval}, we present and discuss the results of an experimental evaluation.
Finally, in Section~\ref{conclusion}, we draw conclusions and discuss possible future research activities.

\section{Pairs of RASP Code and Transformer Weights} \label{dataset}

This section describes how we generate random but functional RASP code and how these RASP programs are then filtered and translated into transformer weights. This development of a novel dataset consisting of pairs of RASP code and the corresponding transformer weights is crucial for training the proposed model.

We use Algorithm~\ref{prog_gen_algorithm} to create random, compileable RASP programs. This algorithm processes the primitives \textit{rasp.tokens} and \textit{rasp.indices} using the RASP functions \textit{Select()}, \textit{Aggregate()}, \textit{SelectorWidth()}, \textit{Map()} and \textit{SequenceMap()} in a way that accounts for the compatibility of functions and the different datatypes. We achieve this by basing the algorithm around a pool of available inputs, which is, in turn, initialized containing only the two primitives \textit{rasp.tokens} and \textit{rasp.indices} (Line 2 of Algorithm~\ref{prog_gen_algorithm})

Algorithm~\ref{prog_gen_algorithm} starts by randomly selecting one of the five RASP functions to use next (Line 6). This random selection is weighted in accordance with how frequently the functions appear in a collection of manually written RASP programs. We then iterate through all of the parameters of the function (Line 7), randomly choosing a variable from the pool of available inputs and test whether it is of the datatype needed for the current parameter~(Line 8 and 9). If we find that one or more of the parameters of the currently selected function cannot be satisfied by any of the currently available inputs, we choose a different function~(Line 11). If this is the case for all of the functions, we restart the generation of the program (Line 12 and 13).

\begin{algorithm}
\begin{scriptsize}
\caption{Random RASP Program Generation}
\label{prog_gen_algorithm}
\begin{algorithmic}[1]
\State $available\_inputs \gets$ Initialize with primitives \textit{rasp.tokens} and \textit{rasp.indices}
\State $available\_functions \gets$ \textit{Select(), Aggregate(), SelectorWidth(), Map(),} and \textit{SequenceMap()}
\State $available\_lambdas \gets$ \textit{2-input lambdas, 1-input lambdas} and \textit{2-input, boolean output lambdas}
\State $phase \gets 0$
\While{not converged to one output}
    \State {Select a function based on its probability from $available\_functions$}
    \For{each parameter of the function}
        \If{appropriate input available for the parameter}
            \State Use inputs from $available\_inputs$ and $available\_lambdas$
        \Else
            \State {Select a different function from $available\_functons$}
            \If{no function is compatible}
                \State Restart the program generation
            \EndIf
        \EndIf
    \EndFor
    \State Add function's output to $available\_inputs$
      \If{phase $= 0$ and termination criterion is met}
        \State $phase \gets 1$
    \EndIf
    \If{phase $= 1$}
        \State Remove used inputs from $available\_inputs$
    \EndIf
\EndWhile
\State \Return The final set of operations forming the RASP program
\end{algorithmic}
\end{scriptsize}
\end{algorithm}

Once a suitable input for all of the function parameters has been found, the output of this function is represented by a new entry in the pool of available inputs (Line 17). Note that some functions, like \textit{Map()} or \textit{Select()}, take lambda functions as some of their inputs. These are not stored in the pool of available inputs but in external lists (Line 3).

In Algorithm~\ref{prog_gen_algorithm}, there are two phases (\textit{phase 0} and \textit{phase 1}). In \textit{phase 0}, we expand the number of available inputs by leaving used variables in the pool of available inputs while adding new variables to this pool. The duration of \textit{phase 0} is controlled via the termination criterion (Line 18). In our implementation, we check whether the length of $(available\_inputs - 2)$ is greater than the duration of \textit{phase 0}. Note that this criterion could be varied to guide the length of the generated programs. In \textit{phase 1}, variables are removed from the available inputs once they are used. Since each function has only one output but typically multiple inputs, this process eventually converges to a single variable, representing the output of the program as a whole.

\begin{figure}
    \centering
    \includegraphics[width=0.76\linewidth]{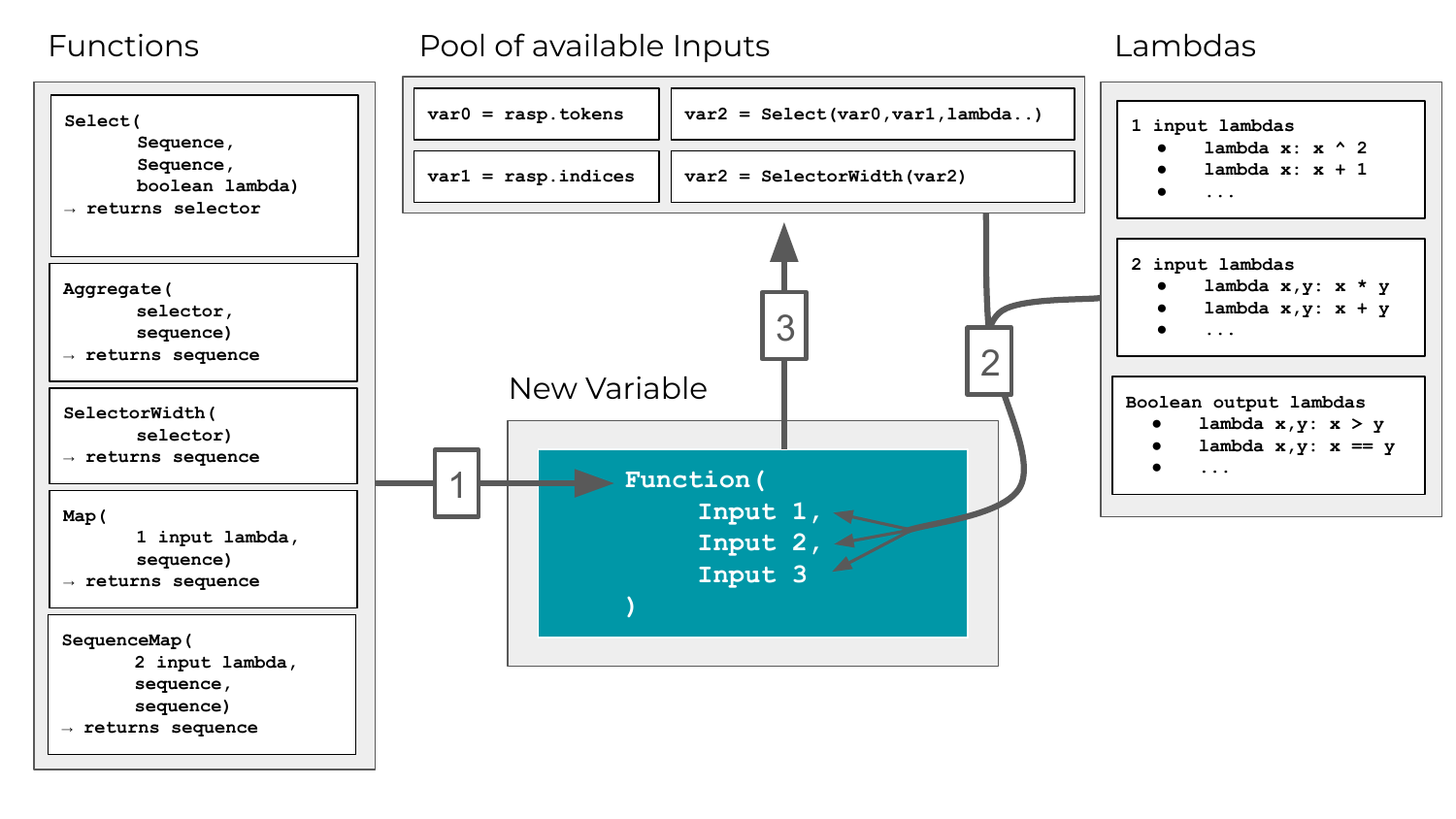}
    \caption{Visualization of the three major steps of Algorithm~\ref{prog_gen_algorithm} that are repeated until the pool of available inputs converges to one entry. 1: Select a function; 2: Fill the function with variables; 3: Add the newly created variable to the pool of available inputs.}
    \label{fig:rasp_gen_algo}
\end{figure}

The output of Algorithm~\ref{prog_gen_algorithm} is a computational graph represented as a list of nodes, each of which has a function and three inputs. Inputs, in turn, are represented by integers, denoting either the node they originate from, a lambda function, or an empty token for functions that take fewer than three inputs. Fig.~\ref{fig:rasp_gen_algo} visually summarizes the process of Algorithm~\ref{prog_gen_algorithm}.

\label{prog_filter}

\begin{figure}
    \centering
    \includegraphics[width=0.8\linewidth]{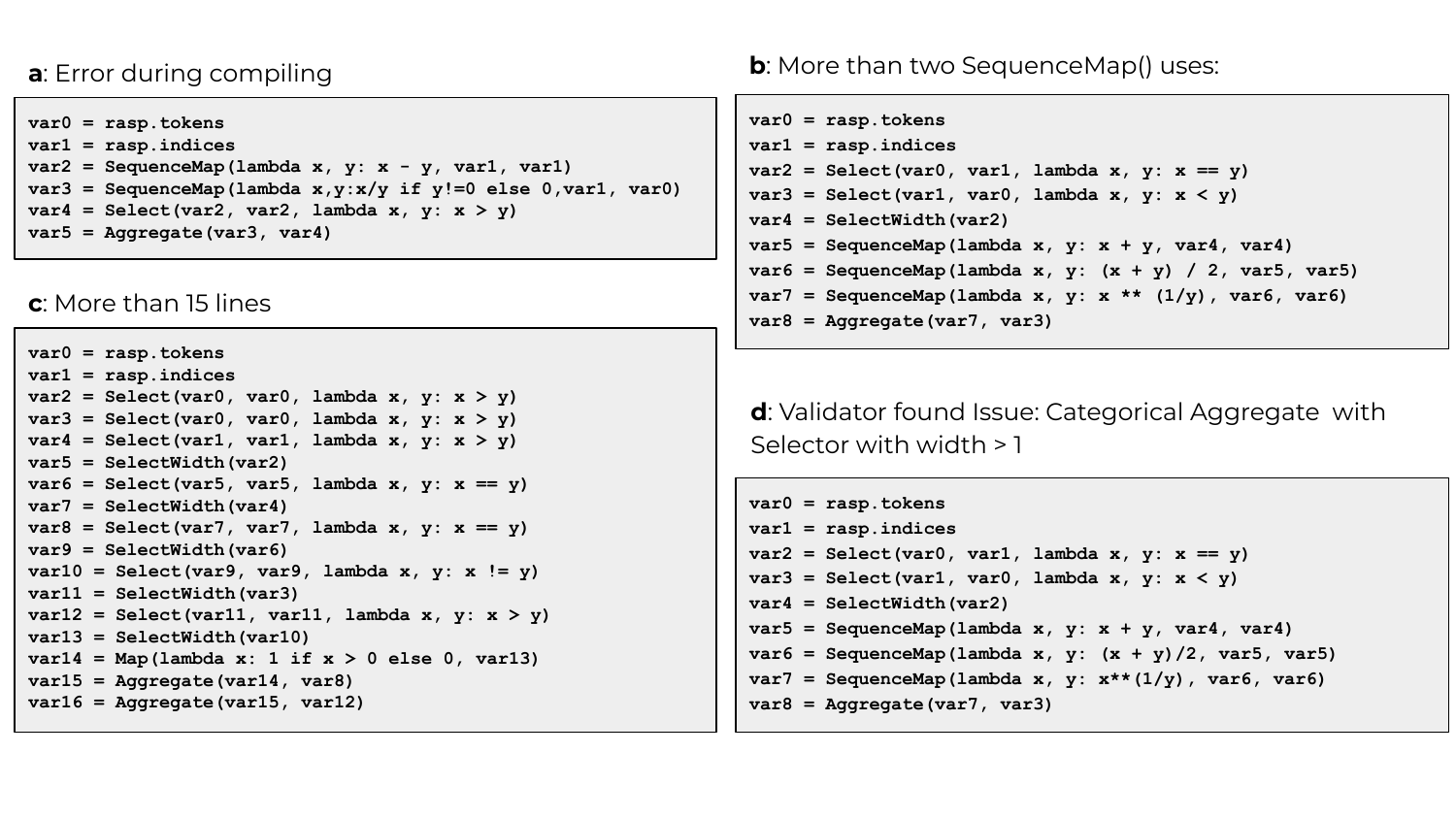}
    \caption{Four examples of generated programs that are rejected}
    \label{fig:rejected_programs}
\end{figure}
As not all programs produced by Algorithm~\ref{prog_gen_algorithm} are suitable for our purpose, we use a rejection sampling strategy. This means we test each program for undesired properties and only keep the ones that pass all the tests. To identify as many flaws or invalidities as possible, we run each program on a set of test inputs and employ the built-in Tracr validator.
Properties that are filtered out relate to some limitations of RASP, like the inability to process float sequences in the \textit{Aggregate()} function and limitations of the Tracr compiler, which does not compile RASP code correctly if it involves the aggregation of non-categorical sequences using a selector with a width that is greater than one. 

In Fig.~\ref{fig:rejected_programs}, we show four examples of generated programs that are rejected due to different reasons:
\begin{enumerate}[label=\textbf{\alph*:}]
\item The Tracr compiler can throw errors on certain valid RASP programs. Such programs are rejected.
\item Multiple uses of the function \textit{SequenceMap()} can cause the compilation to take an unacceptably long time (several minutes rather than less than a second). We deal with this by filtering out programs with more than  two occurrences of \textit{SequenceMap()}.
\item The training set of programs is limited to 15 lines of code which amounts to a maximum of 60 program tokens.
\item The inbuilt Tracr validator filters programs where aggregation of non-categorical values using a selector wider than one occurs.
\end{enumerate}

In total, we produce a dataset that consists of about 533,000 programs and corresponding transformers\footnote{ The generated data is publicly available upon request.}. The produced programs contain between 5 and 12 lines of RASP code which equates to between 20 and 48 tokens. The corresponding transformers consist of between 8 and 42 weight matrices and therefore between 8 and 42 tokens.

To more closely examine the distribution of the programs, we run tests on a subset of 10,000 programs taken from the dataset. For instance, to determine the function\footnote{By function we refer to the algorithm that is implemented by the RASP code.} of each program, we generate a set of 1,000 input sequences. Programs which generate the same output to all of these inputs are then deemed to implement the same function.

During this evaluation, we observe that approximately 60\% of the programs generate the same output for the 1,000 inputs as at least one other program in the dataset. However, the fact that certain functions are equal with respect to their input-output relationship does not necessarily mean that they are internally equal to each other. Actually, we observe that only about 5\% of all programs refer to duplicates (i.e., pairs of programs with identical RASP code strings). However, it is worth mentioning that the small input space of the transformers that are compiled from the RASP programs could mean that two different RASP programs compile to the same transformer weights, leading to some ambiguity in the dataset, where multiple outputs (i.e. programs) would be correct for one input (i.e. weights).


\section{The Transformer Decompiler Method (TraDe)} \label{method}
\subsection{Format of the Model Input}
The formal representation of the RASP code (of the five component functions of the RASP code), into which the Tracr transformer weights are translated, is a crucial step in our procedure. It might be possible to use standard text tokenization~\cite{a2018fast} for this vectorization task. However, this would require the model to learn to distinguish valid RASP code from a very large space of possible outputs. Furthermore, since it is necessary to reverse the vectorization process, the application of commonly used methods for the vectorization of graphs, such as message passing~\cite{vignac2020building}, is not directly applicable to the computational graphs of the RASP program.

\begin{figure}
    \centering
    \includegraphics[width=0.8\linewidth]{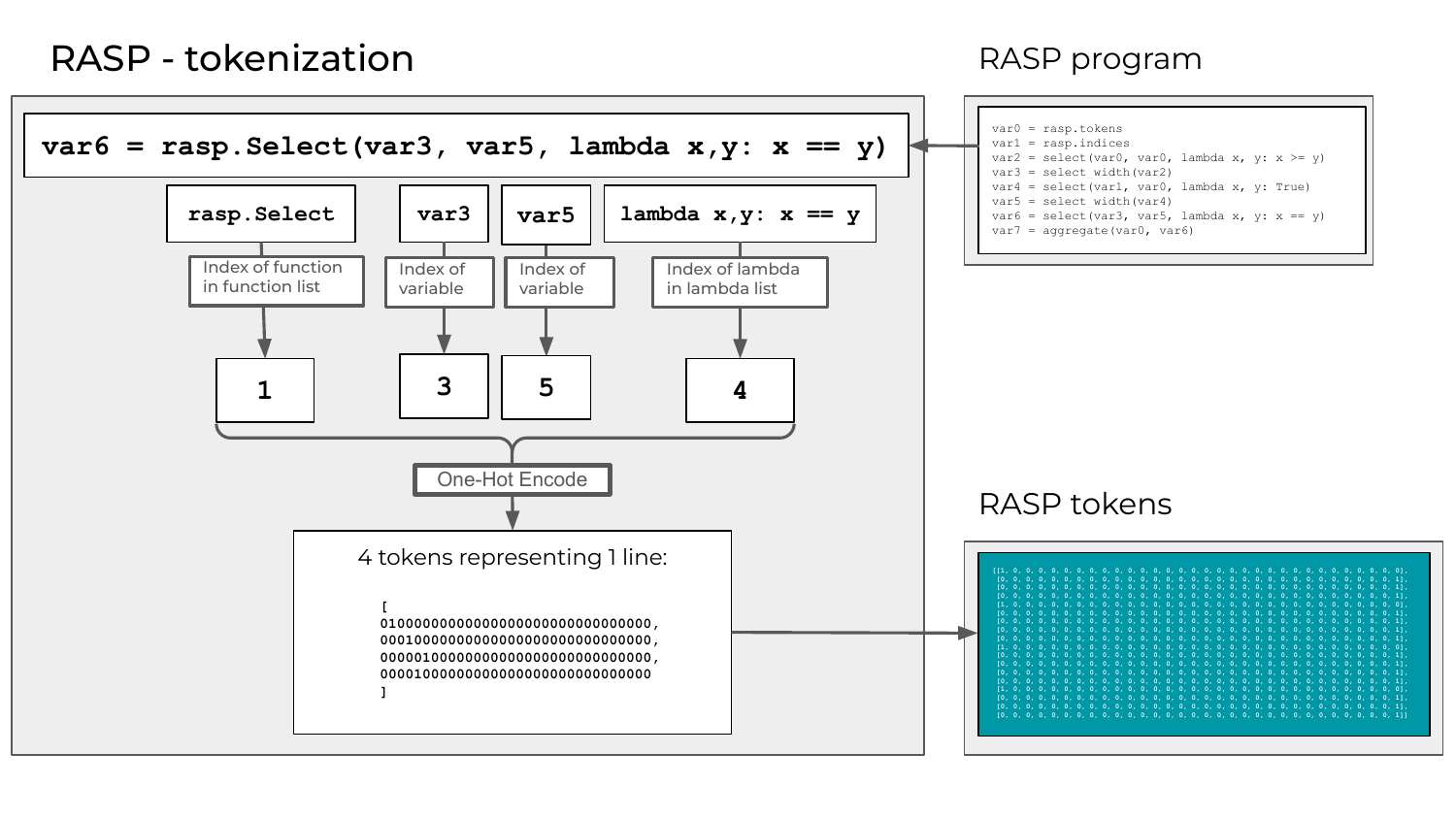}
    \caption{Illustration of the vector representation of the RASP code. It works by dividing each line into four components, which are in turn represented by a one-hot-encoded vector, denoting one of the options for this component. }
    \label{fig:rasp_tokensiation}
\end{figure}

Based on these considerations, we employ a series of one-hot-encoded vectors (four per line of RASP code) to vectorize RASP code. Each series represents a specific part of the line, viz.~the function and the three possible inputs to this function. Depending on the function, these numbers are interpreted differently. For instance, if the first vector of a line represents the function \textit{SequenceMap()}, the next vector will be interpreted as the one-hot-encoded position of the lambda in the list of lambdas that produces one output from two inputs. In Fig.~\ref{fig:rasp_tokensiation}, the process of RASP vectorization is visualized

\begin{figure}
    \centering
    \includegraphics[width=0.8\linewidth]{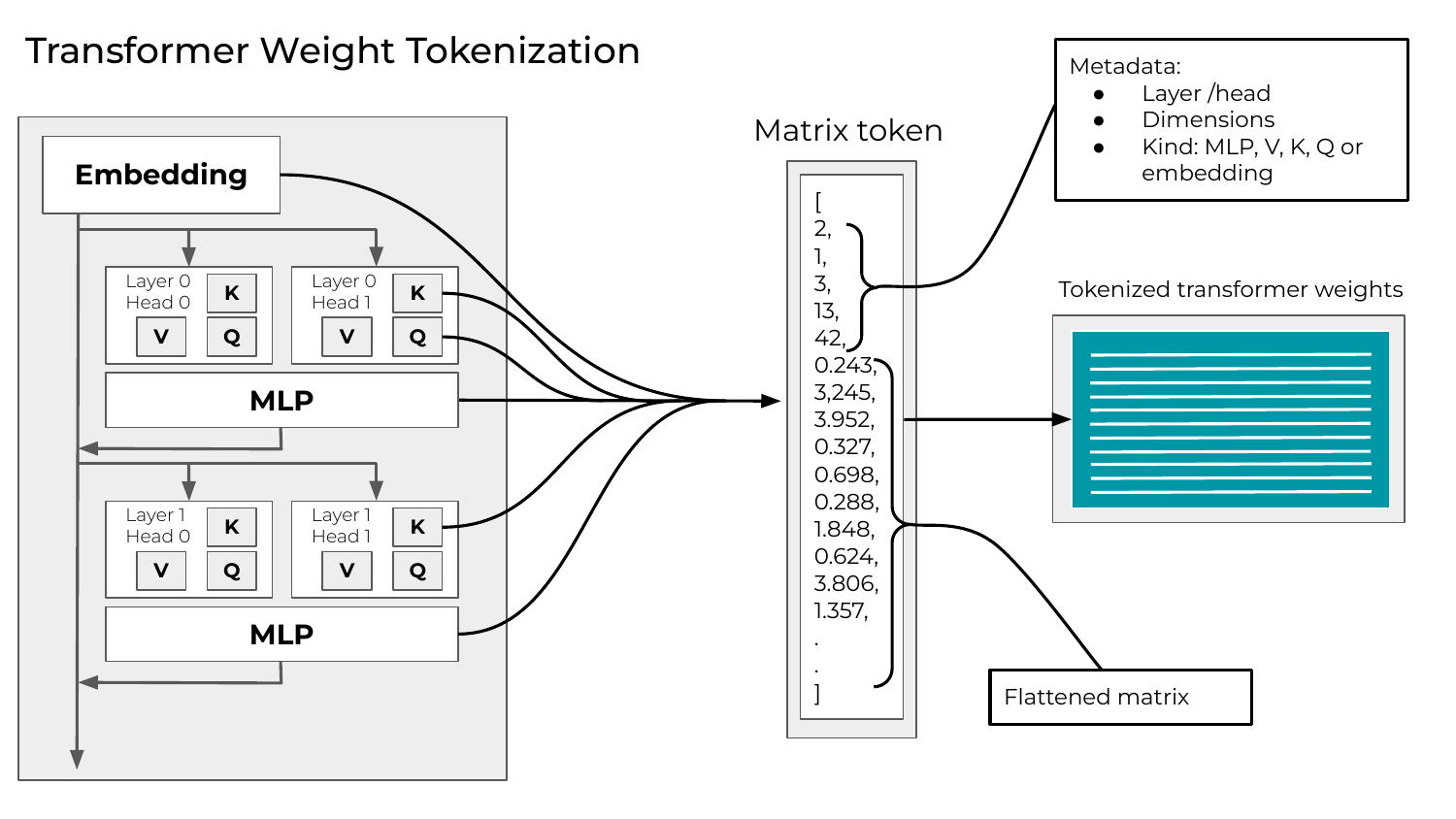}
    \caption{Illustration of the process of translating the weights of a compiled transformer into a set of tokens.}
    \label{fig:tf_tokenisation}
\end{figure}

It is also necessary to bring the weights of the Tracr transformer into a form in which they can be fed into the decompiler. This task is not trivial as transformer weights are not organized sequentially. However, as demonstrated by vision transformers~\cite{dosovitskiy2020image}, the transformer architecture can also process sequences of tokens arranged sequentially, despite their original non-sequential configuration. To retain some of the structure of the transformer, we use a matrix-based tokenization\footnote{We also experiment with other tokenization schemes, such as layer-based tokenization or the naive flattening and partitioning into tokens, of all weights. Matrix-based tokenization emerged as superior in early experiments, though only by a small margin}. That is, each token corresponds to a weight matrix in the compiled transformer. The matrix is flattened and concatenated with a small vector representing some information about the matrix such as shape, position in the transformer (head and layer) and function of the matrix (embedding, MLP, query, key or value).
As the sizes of the matrices vary strongly, the resulting vectors are padded with zeros to achieve equal length.
In Fig.~\ref{fig:tf_tokenisation}, the process of transformer weight tokenization is visualized.

\subsection{Decompiler Model}

Our aim is to solve the problem of translating from the modality of transformer weights to the modality of RASP code. To this end, we adapt an existing system that also translates between modalities; namely the Whisper speech-to-text models by OpenAI~\cite{radford2023robust}. Similar to Whisper, we employ an encoder-decoder transformer architecture. The encoder looks at the model input, in our case the tokenized transformer weights, and guides the decoder which produces the next RASP token based on all of the previously produced ones.


\section{Experimental Evaluation} \label{exp_eval}
To evaluate our model, we generate an additional dataset of 1,000 programs, independent from the original 533K samples used for training.
We optimize the hyperparameters of our model on one Nvidia GTX 3090. Note that the optimization of hyperparameters is based on the \textit{token-accuracy} on a validation set of 28K samples. Token-accuracy refers to the fraction of output tokens that are correct relative to the total number of tokens. That is, every output token of our model is compared with the token at the corresponding position in the program that the transformer weights were compiled from. We are aware that the usefulness of this token-accuracy remains unclear for practical applications. However, it seems plausible that when a majority of the tokens are correct, this allows for the extraction of useful features from the RASP code, like for instance, the causal flow through the network.

The evaluated hyperparameters as well as the best-performing values for all hyperparameters are summarized in Table~\ref{tab:TFDecompilerMetrics}.

\begin{table}
\centering
\label{tab:TFDecompilerMetrics}
\begin{tabular}{l@{\hspace{1em}}r@{\hspace{1em}}r}
\hline
\textbf{Hyperparameter} & \textbf{Explored Range} & \textbf{Optimal Value} \\
\hline
Feature Dimension & \{512, 1024, 2024\} & 512 \\
Heads per Layer (encoder \& decoder) & \{4,6,$\ldots $,16\} & 16 \\
Number of layers: & & \\
\multicolumn{1}{@{\hspace{1em}}l}{Encoder Layers} & \{4,6,$\ldots $16\} & 4 \\
\multicolumn{1}{@{\hspace{1em}}l}{Decoder Layers} & \{4,6,$\ldots $,16\} & 4 \\
Feedforward Dimension & \{1024, 2048\} & 2048 \\
Dropout & \{0.01, 0.1, 0.2, 0.3\}  & 0.2 \\
Input Dimension (\textbf{x}) & & 2000 \\
Output Dimension (\textbf{y}) & & 32 \\
\hline
\end{tabular}
\newline
\caption{Best performing values for all evaluated hyperparameters} 
\end{table}

First, we evaluate the trained decompiler model on our independently generated test set in the non-autoregressive mode, in which it predicts the next token based on the ground-truth prefix.

\begin{figure}
    \centering
    \includegraphics[width=0.6\linewidth]{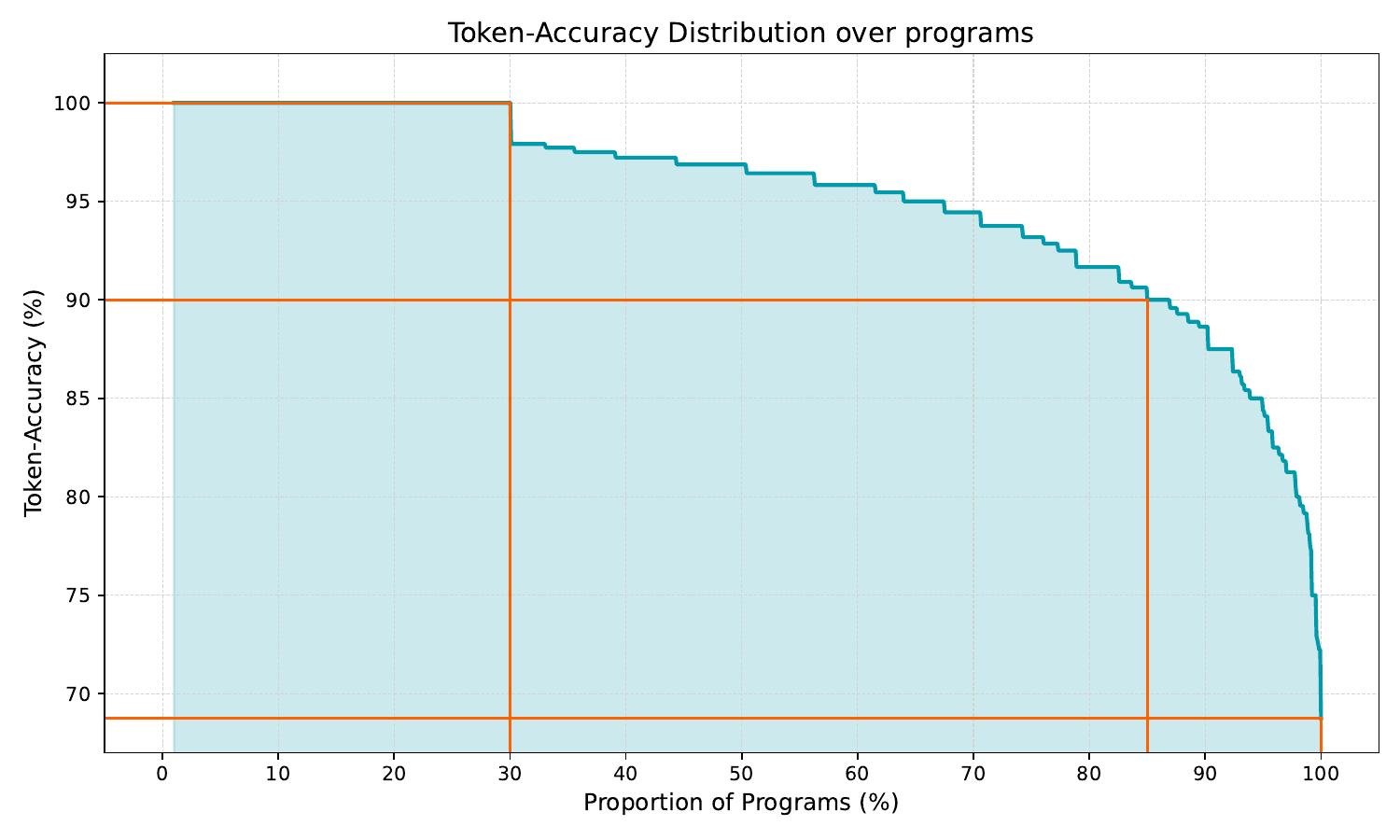}
    \caption{Our model decompiles more than 30\% of programs with 100\% token-accuracy, with the rest of the programs being reproduced with at least 68\% token-accuracy.}
    \label{fig:acc_distribution}
\end{figure}

In Fig.~\ref{fig:acc_distribution} we show the relative proportion of test programs that can be reproduced with a certain token-accuracy. We find that about 30\% of all programs are reproduced identically to the ground truth (i.e., with a token accuracy of 100\%)\footnote{We name the proportion of output programs that are identical to the ground truth \textit{sequence equality}}. Approximately 85\% of all test programs achieve a token-accuracy of 90\% and overall we can report that all programs achieve at least 68\% token-accuracy in this mode.

\begin{figure}
    \centering
    \includegraphics[width=0.8\linewidth, trim={0cm 0cm 0cm 0cm}, clip]{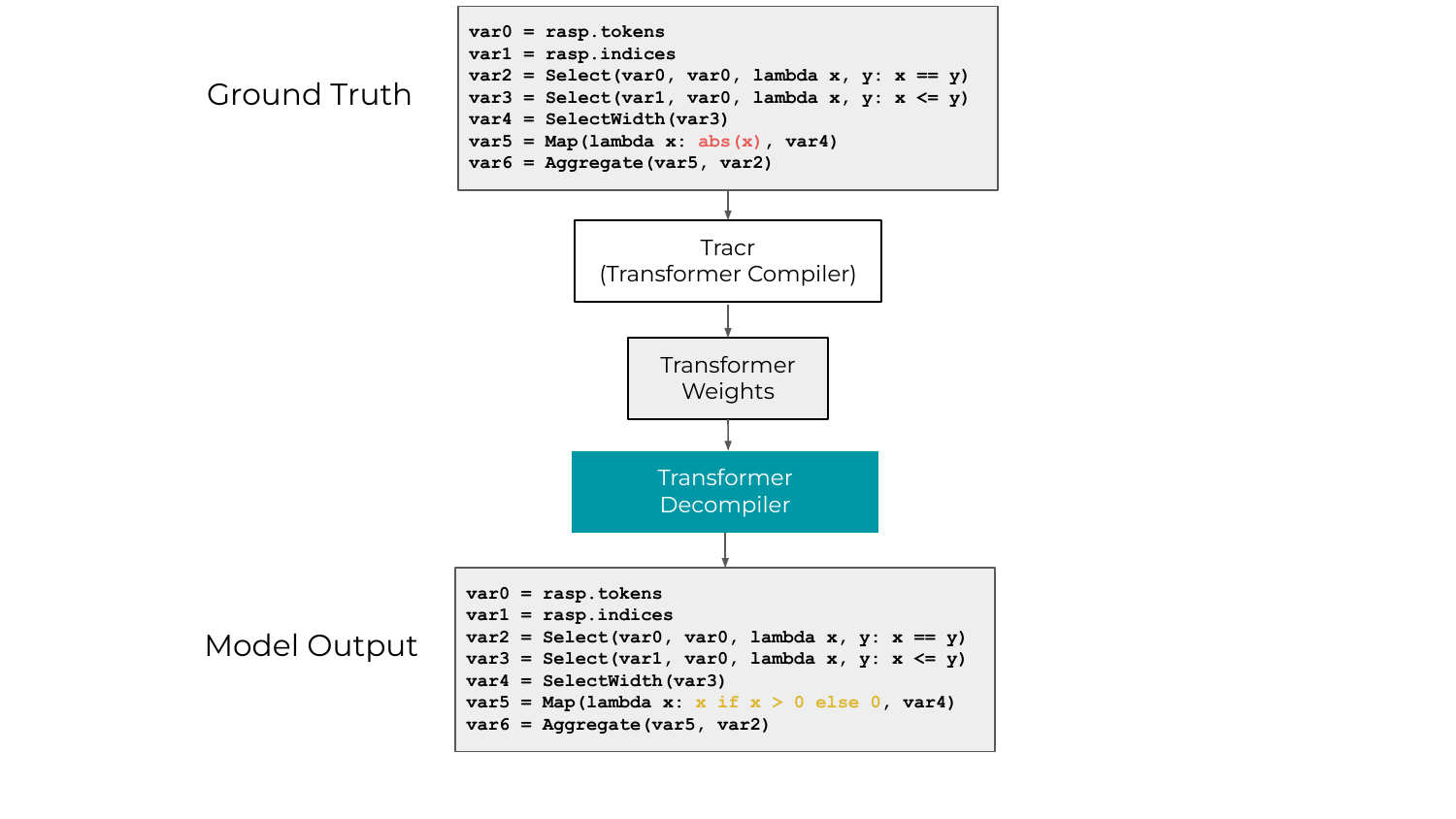}
    \caption{These programs are identified as erroneous reproduction. However, they are functionally equivalent. The \texttt{var4} variable is produced by a \texttt{SelectWidth()} function, which only outputs values equal to or greater than zero. When applied to such values, the functions \texttt{abs(x)} and \texttt{x if x > 0 else 0} are equivalent.
    }
    \label{fig:equiv_input_output}
\end{figure}

Moreover, in this mode we observe that 60\% of the generated code is actually valid and runs without compilation error and 41\% of the output programs are functionally equivalent to the ground truth. This means that they represent the same input-output relations as the RASP code from which the transformer weights are originally compiled (an example of an output that does not match the ground truth but is functionally equivalent to the ground truth is shown in Fig.~\ref{fig:equiv_input_output}).

Next, we test the model in autoregressive mode, in which it produces each token based on the tokens it previously produced. In this mode, the relative number of programs that achieve 100\% token accuracy (i.e., sequence equality) drops from 30\% to 26\%. However, 91\% of the outputs are now compilable (rather than 60\% as achieved in the non-autoregressive mode). Remarkably, 73\% (rather than 41\%) of the output programs are functionally equivalent to the ground truth RASP program that the transformer was compiled from. This could be explained by the model not seeing its previous outputs, but only the beginning of the ground-truth-program in the non-autoregressive mode. Seeing its own outputs for previous lines might allow the model to act according to the decisions it made earlier in the generation process.
 
\begin{figure}
    \centering
    \includegraphics[width=0.8\linewidth]{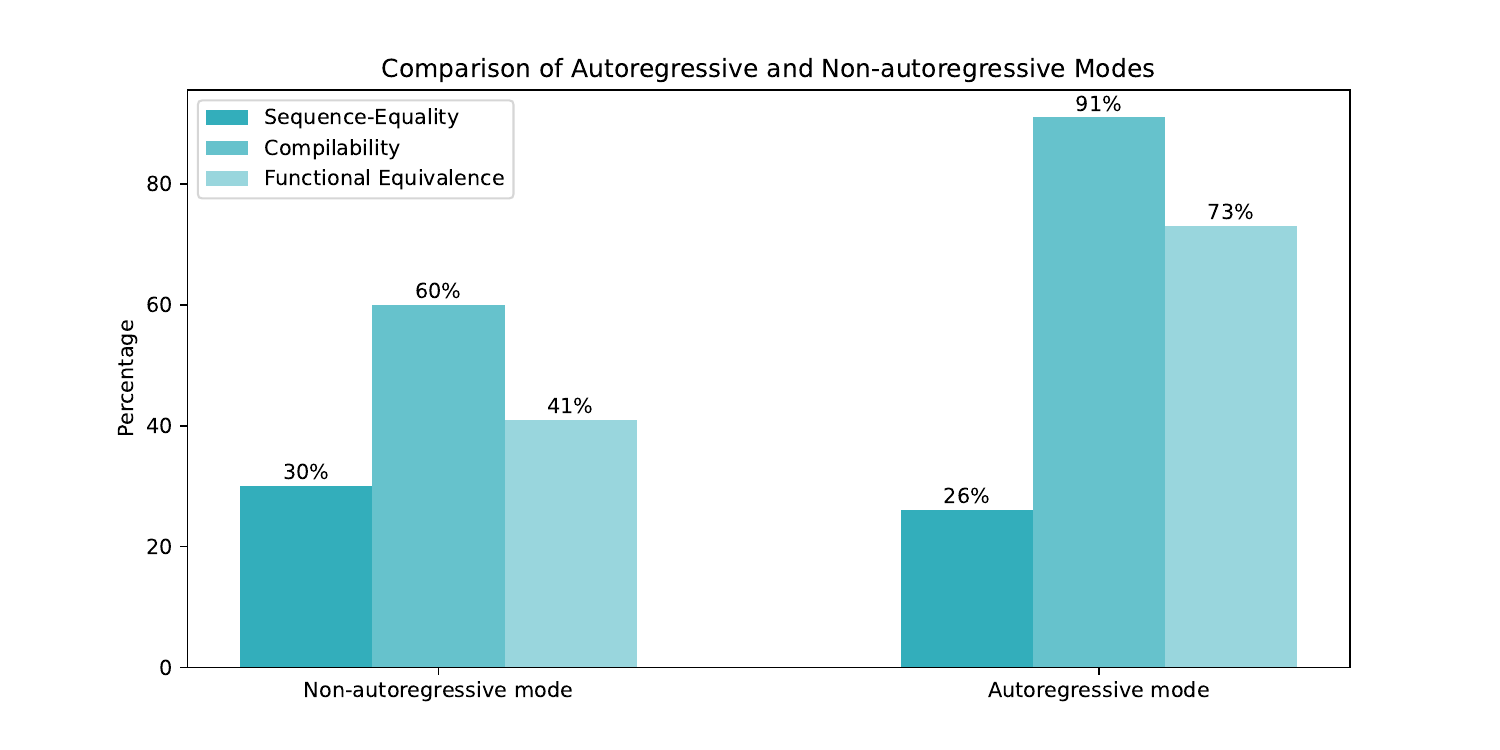}
    \caption{Comparison of autoregressive and non-autoregressive mode across sequence equivalence to the ground truth, compilability and functional equivalence to the ground truth.}
    \label{fig:ar_vs_nonar}
\end{figure}

The relative amount of outputs that achieve sequence equality,  compilability, as well as functional equivalence are summarized in Fig.~\ref{fig:ar_vs_nonar} for both modes (non-autoregressive and autoregressive). 

\section{Conclusion} \label{conclusion}
To date, there are no general solutions available for the automatic interpretation of neural network models. In the present paper, we suggest that the Transformer Decompiler Method (TraDe) could serve as an approach to address this issue. In an empirical evaluation we show that TraDe enables a model to interpret the weights of other, smaller transformer neural networks and translate them into a more human-readable modality with useful accuracy. Our work thus represents a significant step towards an end-to-end framework for better interpretability.

However, there are still many limitations standing in the way of any practical application of the proposed concept. For instance, the transformer weights resulting from the Tracr compilation process are very different to transformer weights resulting from an optimization using stochastic gradient descent. The former is very sparse (except for certain structured elements), while the latter is very unstructured and dense. Moreover, the best-performing variant of the decompiler model is about three orders of magnitude larger than the models it is capable of decompiling (in terms of parameters). If this ratio is not significantly reduced, the application to modern, large transformer models will remain impossible. Lastly, though the step from weight matrices to RASP code is a large improvement in terms of interpretability, it would be wrong to call the produced RASP programs human-readable. Even with the simple 5 to 12 lines of RASP code, it can take some minutes for a human to determine what sequence operation the algorithm implements.

Possible future work is concerned with reducing (or eliminating) the above-mentioned limitations.
For instance, by taking the current decompiler model on a set of compressed Tracr models, it might be possible to adapt the decompiling Tracr transformer weights to the task of decompiling learned transformer weights (compressed Tracr models contain matrices that are trained with gradient descent, and thus might more closely reflect realistic weights). Another option might be to reduce the task of the decompiler from the reproduction of all RASP code to the detection of certain features, like backdoors~\cite{chen2017targeted} or deceptive tendencies~\cite{hagendorff2023deception,scheurer2024large}, or to the analysis of a sub-component of the transformer~\cite{langosco2024meta}.
Last but not least, it could also be interesting to see which weight matrix is attended to the most when producing a certain piece of RASP code. 


\begin{credits}
\subsubsection{\ackname} This study was funded by the \textit{Hasler Foundation Switzerland} (grant number 23085).

\subsubsection{\discintname}
The authors have no competing interests to declare that are
relevant to the content of this article. 
\end{credits}

\bibliographystyle{unsrt}
\bibliography{lib}

\end{document}